\documentclass{article}

\usepackage{arxiv}
\usepackage{graphicx}

\usepackage[utf8]{inputenc} 
\usepackage[T1]{fontenc}    
\usepackage{hyperref}       
\usepackage{url}            
\usepackage{booktabs}       
\usepackage{amsfonts}       
\usepackage{nicefrac}       
\usepackage{microtype}      
\usepackage{lipsum}

\title{DeepPFCN: Deep Parallel Feature Consensus Network for Person Re-Identification}

\author{
  Shubham Kumar Singh \\
  Department of CSE\\
  IIT Gandhinagar\\
  Gujarat, INDIA \\
  \texttt{shubham.singh@iitgn.ac.in} \\
   \And
   Krishna P Miyapuram \\
  Department of CS \& CSE\\
  IIT Gandhinagar\\
  Gujarat, INDIA \\
  \texttt{kprasad@iitgn.ac.in} \\
   \And
  Shanmuganathan Raman \\
  Department of EE \& CSE \\
  IIT Gandhinagar\\
  Gujarat, INDIA \\
  \texttt{shanmuga@iitgn.ac.in} \\
}

\begin{document}
\maketitle

\begin{abstract}
Person re-identification aims to associate images of the same person over multiple non-overlapping camera views at different times. Depending on the human operator, manual re-identification in large camera networks is highly time consuming and erroneous. Automated person re-identification is required due to the extensive quantity of visual data produced by rapid inflation of large scale distributed multi-camera systems. The state-of-the-art works focus on learning and factorize person appearance features into latent discriminative factors at multiple semantic levels. We propose Deep Parallel Feature Consensus Network (DeepPFCN), a novel network architecture that learns multi-scale person appearance features using convolutional neural networks. This model factorizes the visual appearance of a person into latent discriminative factors at multiple semantic levels. Finally consensus is built. The feature representations learned by DeepPFCN are more robust for the person re-identification task, as we learn discriminative scale-specific features and maximize multi-scale feature fusion selections in multi-scale image inputs. We further exploit average and max pooling in separate scale for person-specific task to discriminate features  globally and locally. We demonstrate the re-identification advantages of the proposed DeepPFCN model over the state-of-the-art re-identification methods on three benchmark datasets - Market1501, DukeMTMCreID, and CUHK03. We have achieved mAP results of 75.8\%, 64.3\%, and 52.6\% respectively on these benchmark datasets.
\end{abstract}


\section{Introduction}
\label{sec:intro}

Person re-identification detects whether a person of interest has been observed in another place (time) by a different camera ~\cite{zheng2015scalable}. In many scenarios, people appearing in one camera do not necessarily appear in another camera and sometimes the camera view may include people that have never appeared in any other camera  as well.  Therefore, it is better to treat person re-identification as a verification problem ~\cite{Gong2014person}. Person re-identification is a naturally challenging task because correctly matching two images of the same person is difficult under extensive appearance changes, such as human pose, illumination, occlusion, background clutter, (non)uniform clothing, and camera view-angle \cite{Gong2014person}. It has applications in tracking a particular person across these cameras, tracking the trajectory of a person, real time surveillance, and forensic and security applications.

  In this paper, we propose Deep Parallel Feature Consensus Net (DeepPFCN) a novel network architecture that learns multi-scale person appearance features using convolutional neural networks (CNN) \cite{chen2017person} and factorizes the visual appearance of a person into latent discriminative factors at multiple semantic levels \cite{chang2018multi}. DeepPFCN is a combination of the above mentioned two architectures, which are orthogonal to each other as mentioned in \cite{chang2018multi}. DeepPFCN focuses on fusing both automated discovery of latent appearance factors and fusing image resolutions. DeepPFCN deploys a multi-loss concurrent supervision mechanism. This allows enforcing and improving scale-specific feature learning.
 
  DeepPFCN is evaluated on three person re-identification benchmark datasets - Market1501 \cite{zheng2015scalable}, DukeMTMCreID \cite{zheng2017unlabeled}, CUHK03 \cite{li2014deepreid}. Extensive experiments and ablation study have been conducted on these datasets. In particular, we achieve the mAP scores of 75.8\%, 64.3\%, and 52.6\% on the above mentioned benchmark datasets, which is observed to be better than the state-of-the-art methods by 1.5\%, 1.5\% and 3.4\%, respectively.  

 This paper is organized as follows. In section \ref{sec:architecture}, we describe the components of DeepPFCN architecture - (a) modified multi-level factor net as base model and (b) multi-scale consensus learning with back-propagation. We combine these methods, which achieves the state-of-the-art performance for person re-identification. In section \ref{sec:experiment}, we work with the following datasets: Market1501 \cite{zheng2015scalable}, DukeMTMCreID  \cite{zheng2017unlabeled} and CUHK03 \cite{li2014deepreid} and explain the evaluation metrics, data augmentation, training and evaluation in detail. In section \ref{sec:result}, we describe the experiments and present the ablation study conducted on these datasets. Finally, section \ref{sec:conclusion} concludes the paper.

\section{Related Work}
\label{sec:work}
 
 Person re-identification is considered as an important task in the field of computer vision. Several researchers have attempted to find effective and efficient person re-identification solution. Existing methods can be either traditional methods \cite{chen2015mirror,koestinger2012large,li2015multi,liao2015person} or deep learning based methods \cite{ahmed2015improved,chang2018multi,chen2017person,li2014deepreid,li2017person}. Recently, deep learning based person re-identification models have obtained excellent performance. In practice, two types of models are used for person re-identification: verification model and identification model.  
 
  For verification model, a Siamese neural network \cite{koch2015siamese} or triplet loss is exploited to make feature vector similar for a pair of images with same identity and dissimilar for different identities \cite{ahmed2015improved,deng2009imagenet,chang2018multi,li2014deepreid}. For the identification model in \cite{xiao2016learning}, a discriminative representation of the given input image is learned and it is generally observed to perform better than the verification model. However, existing re-identification methods typically consider only one resolution scale of person appearance which is potentially not considered as an useful information on another scale and it also loses the correlated complementary advantage across different scales appearance. 

 Multi-Level Factorization Net (MLFN) \cite{chang2018multi} come up with a discriminative latent factor with no secondary supervision. Multi-level factors are shared by all network blocks rather than overloading the final layer. However, all the existing methods with one resolution scale of person appearance does not only drop the potential useful information of other scales, but also lose the correlated complementary advantage across appearance scale. Deep Pyramidal Feature Learning (DPFL) \cite{chen2017person} deploys a multi-loss concurrency for improving scale-specific feature individuality learning. Softmax classification loss is used in this method for reducing model training complexity and improve the model learning scalability when a large, different scale data is provided. 

 Our approach is aimed to combine both the orthogonal architectures into one single model called DeepPFCN.  DeepPFCN  not only learns discriminative scale-specific features and maximize multi-scale feature fusion selections in multi-scale image  inputs, but also exploits  average  and  max  pooling in  separate  scales  for  person-specific task to discriminate features globally and locally.   

 \textbf{Our Contributions} are as follows. (1) DeepPFCN is self-reliant to discover discriminative and view-invariant appearance features at multiple scales without secondary supervision. (2) We exploit average and max pooling in separate scales for person-specific task to  discriminate features globally and locally. We obtain the  state-of-the-art results on three large person re-identification benchmark datasets - Market1501 \cite{zheng2015scalable}, DukeMTMCreID  \cite{zheng2017unlabeled}, and CUHK03 \cite{li2014deepreid}.

\section{ Model Architecture}
\label{sec:architecture}

This section describes the problem statement and structure of Deep Parallel Feature Consensus Network (DeepPFCN) architecture. DeepPFCN architecture has the following components: (a) modified multi-level  factorization network as base model and (b) multi-scale consensus learning network with back-propagation. We combine these components, which achieves new state-of-the-art results in person re-identification.

\subsection{Problem Statement}
Consider a set of $n$ training images $I = \left \{I_i \right \}^n_{i=1}$ with the corresponding identity class labels $\gamma = \left \{ y_i \right \}^n_{i=1} $.  These training images capture the visual appearance and the variation of $n_{id}$  (where $y_i \in [ 1, \ldots$ , $n_{id} ]$ ) different people under multiple non-overlapping camera views. A re-identification model needs to learn from these image-identity corresponding relations and use the learned knowledge to recognize other unseen person identities. We have formulated a DeepPFCN model that aims to extract discriminative appearance information about person identity from multiple resolution scales  under significant viewing condition changes across distinct locations.

\subsection{Parallel Feature Consensus Learning }
The overall network design of the proposed DeepPFCL model is depicted in Figure \ref{fig2}. This DeepPFCL model has ($m + 1$) feed-forward sub-network branches: (1) $m$ branches of scale-specific sub-networks with an identical structure for learning the most discriminative visual features for each individual pyramid scale of person bounding box images, (2) One fusion branch responsible for learning the discriminative feature selection and optimal integration of $m$ scale-specific representations of the same images and average and max pooling in separate scale for person-specific task to discriminate features globally and locally. We describe the architecture components below: (i) modified multi-level factor net as base model and (ii) multi-scale consensus learning with back-propagation.

\begin{figure}[t]
\includegraphics[width=\textwidth]{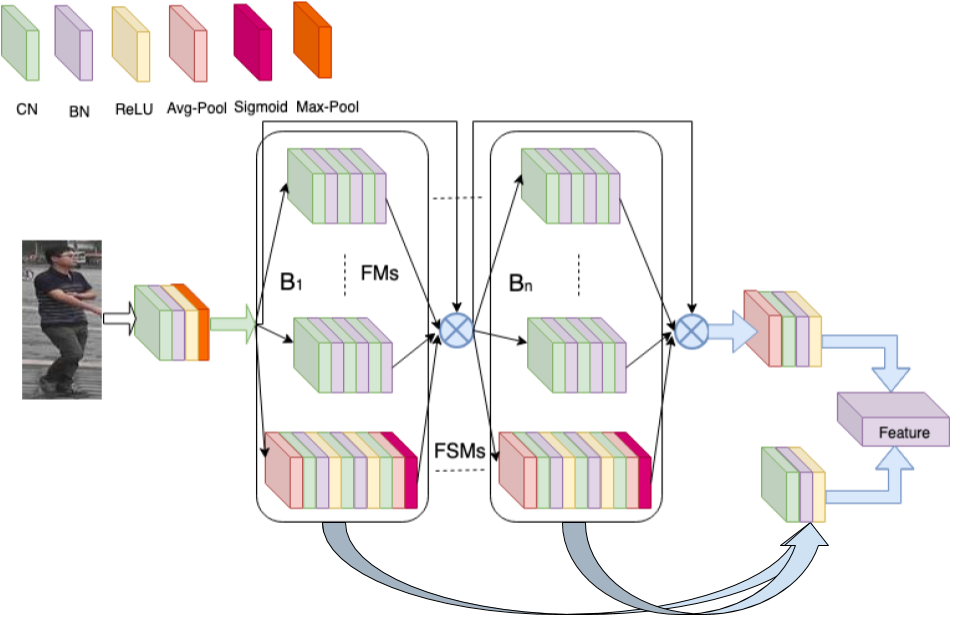}
\caption{Illustration of Multi-Level Factorisation Network (MLFN)
Architecture.\cite{chang2018multi}} \label{fig1}
\end{figure}

 \textbf{Multi-Level Factorisation Network}:  The work by \cite{chang2018multi} aims to automatically discover latent discriminative factors at multiple semantic levels and dynamically identify their presence in each input image. As shown in Figure \ref{fig1}, $N$  blocks are stacked to model $N$ semantic levels. For person re-identification task, within each block, 32 Factor Modules (FMs) and 1 Factor Selection Module (FSM) are used and 16 blocks are let up in MLFN. Let $B_n$ denotes the $n$th block, $n$ $\in$ {$1$,$\ldots$, $N$} and within each $B_n$, there are two key components: FMs and a FSM. FMs help the model to learn a latent factor at the corresponding level indexed by $n$. FSM handles the case where multiple discriminative latent factors are required simultaneously to explain the visual appearance of the input image. MLFN fuses the deep feature computed from the final block $B_N$ and the Factor Signature (FS). The final output representation of MLFN is computed by averaging the two projected features. A final fully connected layer is added that projects it to a dimension matching the number of training classes.

 \textbf{Multi-Scale Consensus Learning}:
We perform multi-scale consensus learning on person identity classes from $m$ scale-specific branches. In PFCL instantiation by modified MLFN, we achieve the feature fusion by the operation of concatenation. The loss at final layer is added with scale-specific loss and back-propagated in each scale respective. The proposed DeepPFCN model uses a kind of knowledge transfer dynamically in an interactive manner. The entire feature learning by multi-scale person identity consensus is a close-loop. We further propagate the consensus as extra feedback information to regularize the batch learning of all scale-specific branches concurrently as shown in Figure \ref{fig2}. For person re-identification tasks, in single scale MLFN 32-D FSM output vector is generated and FS becomes 32 FM $\times$ 16 blocks = 512. As a result, each MLFN output is combination of FM's and FMS's output. The final feature dimension of parallel feature consensus net is set to twice of the MLFN feature vector.

\begin{table}[t]
\begin{center}\begin{tabular}{|l|l|l|}
\hline
Dataset & Market1501 & DukeMTMCreID \\ 
\hline
Metric (\%)  & Rank-1 \hfill mAP & Rank-1 \hfill mAP \\ 
\hline
(192x96) & 88.8 \hfill 73.4 & 81.4 \hfill 62.0 \\ 
\hline
(256x128) & 90.4 \hfill 75.8 & 82.0 \hfill 63.4 \\ 
\hline
(384x192) & 91.1  \hfill 75.9 & 81.1 \hfill 63.9 \\ 
\hline
\end{tabular}
\end{center}
\caption{Comparison of Multi-scale resolution to MLFN model.}
\label{table:1}
\end{table}

We perform an experiment on three different scale resolution as $192 \times 96$, $256 \times 128$, and $384 \times 192$ on  the  state-of-the-art for Market1501 and DukeMTMCreID datasets as shown in Table \ref{table:1}. We exploit m = 2 resolution scale for input image: $384 \times 192$ and $256 \times 128$, based on the better results obtained on the respective datasets.

\begin{figure}[t]
\includegraphics[width=\textwidth]{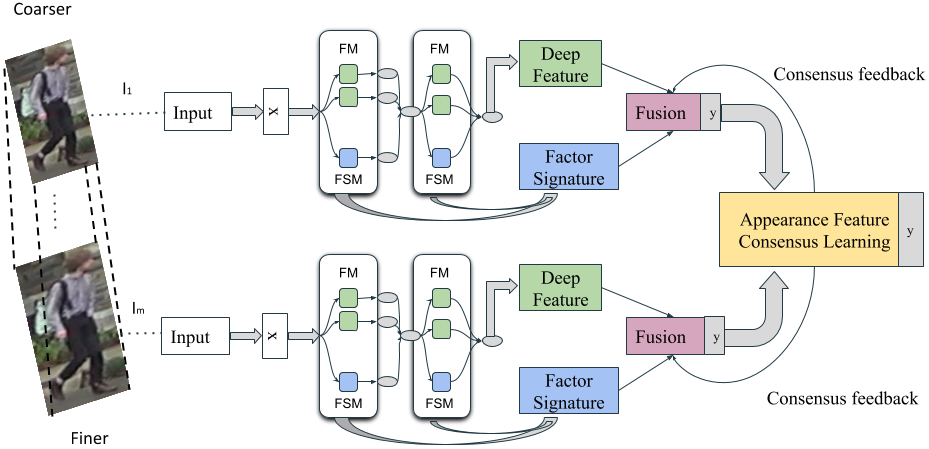}
\caption{Overview of the proposed Deep Parallel Feature Consensus Learning (DeepPFCL). The feature representations learned by DeepPFCN are more robust as: (a) we learn discriminative scale-specific features and maximize multi-scale feature fusion selections in multi-scale image inputs and (b) we exploit average and max pooling in separate scales for person-specific task to  discriminate features globally and locally.} \label{fig2}
\end{figure}

\section{ Experiments}
\label{sec:experiment}

\subsection{Datasets}

The following three person re-identification benchmarks -  Market1501 \cite{zheng2015scalable}, Duke MTMCreID \cite{zheng2017unlabeled},  CUHK03 \cite{li2014deepreid}. Each of the above mentioned individual datasets are used for testing. \textbf{Market1501} \cite{zheng2015scalable} has 12,936 training, 3368 query and 15913 gallery images with 1501 identities in total from 6 cameras. Deformable Part Model(DPM) \cite{felzenszwalb2010object} is used for person detection. \textbf{DukeMTMCreID} \cite{zheng2017unlabeled} comprises of 16522 training, 2228 query and 17661 gallery images with 1404 identities in total from 8 cameras. Manually labelled pedestrian boxes are provided for person detection. \textbf{CUHK03} \cite{li2014deepreid} contains 7365 training, 1400 query and 5332 gallery images with 1467 identities in total from 2 cameras. Both manually labelled and DPM detected person bounding boxes are provided. 

\subsection{Evaluation metrics}
Cumulated Matching Characterstics (CMC) curve and mean Average Precision (mAP) as suggested in \cite{zheng2015scalable} are used to evaluate the performance of re-identification methods. We only report the CMC and mAP for rank-1 in tables rather than plotting the actual curves. All results described in this paper is under single query and does not use post-processing re-ranking by \cite{zhong2017re}. 

\subsection{Data Augmentation}
 We use five data augmentation techniques during training, such as, random cropping \cite{krizhevsky2012imagenet}, random erasing \cite{zhong2017random}, flipping \cite{simonyan2014very}, color jitter, and color augmentation. Random cropping \cite{krizhevsky2012imagenet} is used to reduce the contribution of the background. Random erasing \cite{zhong2017random}, which randomly select a rectangle region in an image and erases its pixels with random values. Left-right flip \cite{simonyan2014very} augmentation is used so that an image should be equally recognizable as its mirror image. Color jitter randomly changes brightness, contrast, and saturation. Color augmentation randomly alters the intensities of RGB channels. No data augmentation is used for testing.

\subsection{Training and Evaluation}
All the person re-identification models are fine tuned on ImageNet \cite{deng2009imagenet} pre-trained networks. The Adam \cite{kingma2014and} optimizer is used with a mini-batch size of 16. We use initial learning rate of 0.0005 in CUHK03 with setting 2 \cite{zhong2017re}  and 0.0003 for the remaining two datasets. Momentum terms are $\beta_1$ = 0.5, $\beta_2$ = 0.999. The number of training iterations is 80 epochs for all the person re-identification datasets. Cross-Entropy loss is used during training.

\section{Result}
\label{sec:result}

\subsection{Comparison with the State-of-the-art Methods}

\begin{figure}[t]
\includegraphics[width=\textwidth]{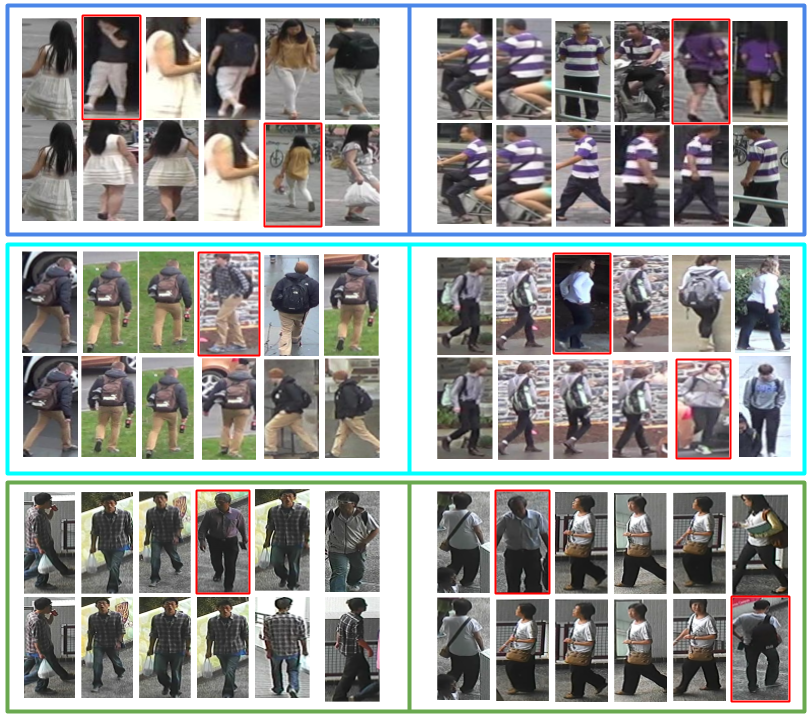}
\caption{Visualising person re-identification performance by MLFN \cite{chang2018multi} (Upper row) and DeepPFCN (lower row). For three groups of images from Market1501 \cite{zheng2015scalable}, DukeMTMCreID \cite{zheng2017unlabeled} and CUHK03 \cite{li2014deepreid} benchmark datasets respectively, first image of each example shows the probe person image, followed by upto Rank-5 gallery images by respective methods with red box indicating false matches.} \label{fig3}
\end{figure}

\textbf{Results on Market1501} Comparison between DeepPFCN and the state-of-the-art  approaches on Market1501 \cite{zheng2015scalable} are shown in Table \ref{table:2}. The results show that our DeepPFCN achieves the best performance with the mAP of 75.8\% and Rank-1 accuracy 90.6 \%, which outperform all the existing works by more than 1.5\% and 0.6\%, respectively. No post-processing operation (eg. the re-rank algorithm \cite {zhong2017re}) is used here. Results on Market1501 dataset can be observed in the first two rows of Figure \ref{fig3}.

\noindent \textbf{Results on DukeMTMCreID} Table \ref{table:2} shows the results on DukeMTMCreID \cite{zheng2017unlabeled}. This dataset contains person boundary box of varying size across different camera view. This challenge is best suited for our multi-scale architecture. Without any post-processing, DeepPFCN achieves the result of mAP, 64.3 and Rank-1 accuracy as 82.1 \%. This is better than the state-of-the-art methods by 1.5\% and 1.1\% respectively. Results can be shown on DukeMTMCreID dataset corresponding to third and fourth rows of Figure \ref{fig3} with two distinct example.

 \noindent \textbf{Results on CUHK03} Person re-identification results on CUHK03 \cite{li2014deepreid} are given in Table \ref{table:2} with setting 2 when detected person boundary boxes are used for both training and testing. Our DeepPFCN performs better in terms of the mAP (52.6\%) and Rank-1 accuracy (56.7 \%) on this dataset, which surpasses all existing works more than 3.4\% and 2.0 \% respectively. We can see the results on DukeMTMCreID dataset corresponding to fifth and sixth rows of Figure \ref{fig3} with two distinct example.

\begin{table}[t]
\begin{center}
\begin{tabular}{|l|c|c|c|}
\hline
Method &  Market1501 & DukeMTMC & CUHK03 \\
\hline
&R1 \hfill mAP &R1 \hfill mAP& R1 \hfill mAP\\
\hline
SVDNet \cite{sun2017svdnet} & 82.3 \hfill 62.1 & 76.7 \hfill 56.8& 41.5 \hfill 37.3 \\
ACRN \cite{schumann2017person}& 83.6\hfill62.6 & 72.6\hfill52.0 & -\hfill- \\
JLML \cite{li2017person}& 83.9\hfill64.4 & -\hfill- & -\hfill- \\
LSRO \cite{zheng2017unlabeled}  & 84.0\hfill66.1 & -\hfill- & -\hfill- \\
DPFL \cite{chen2017person} & 88.9\hfill73.1 & 79.2\hfill60.6 & 40.7\hfill37.0 \\
MLFN \cite{chang2018multi}&  90.0\hfill74.3 & 81.0\hfill62.8 & 54.7\hfill49.2 \\
\textbf{DeepPFCN (Ours)} & \textbf{90.6}\hfill\textbf{75.8} & \textbf{82.1}\hfill \textbf{64.3} & \textbf{56.7}\hfill\textbf{52.6} \\
\hline
\end{tabular}
\end{center}
\caption{Comparison of the proposed method with the state-of-the-art on Market1501, DukeMTMCreID and CUHK03.}
\label{table:2}
\end{table}

\subsection{Ablation Study}

Recall that our DeepPFCN determines the scale-specific feature learning and optimize discriminative feature selection from multi-scale representation of re-identification, by aggregating modified MLFN with identical structures within each scale. \textbf{DeepPFCN}: Multi-Scale full model with average  and  max  pooling use in separate  scale. \textbf{MLFN}: Single scale DeepPFCN using average polling \cite{chang2018multi}. \textbf{MLFN-Fusion}: MLFN using dynamic factor selection without FS feature. \textbf{ReNeXt}: All FMs are always active and FSMs are eliminated from MLFN so then it becomes ResNeXt \cite{xie2017aggregated}. \textbf{ResNet}: When the sub-network at each level are replaced with one large holistic residual module which is ResNet \cite{he2016identity}.   

\begin{table}[t]
\begin{center}
\begin{tabular}{|l||c||c||c|}
\hline
Datasets &  Market1501 & DukeMTMC & CUHK03 \\
\hline
Methods &R1 \hfill mAP &R1 \hfill mAP& R1 \hfill mAP\\
\hline
ResNet \cite{he2016identity} &  84.3 \hfill 66.0  &  71.6\hfill48.6 &  41.7\hfill 37.9  \\
ResNeXt \cite{xie2017aggregated} & 88.0 \hfill 69.8 & 75.7 \hfill 54.1 & 43.8\hfill 38.7 \\
MLFN-Fusion\cite{chang2018multi}& 87.9 \hfill 70.8 & 78.7 \hfill 58.4 & 47.1 \hfill 42.5  \\
MLFN \cite{chang2018multi}& 90.0 \hfill 74.3 & 81.0 \hfill 62.8 & 54.7 \hfill 49.2  \\
\textbf{DeepPFCN} & \textbf{90.6} \hfill \textbf{75.8} & \textbf{82.1} \hfill \textbf{64.3} & \textbf{56.7} \hfill \textbf{52.6}  \\
\hline
\end{tabular}
\end{center}
\caption{Ablation Results on three Person re-identification datasets. CUHK03 results were obtained under Setting 2.}
\label{table:3}
\end{table}

 A comparison of these models on all three person re-identification dataset is shown in Table \ref{table:3}. We can see that DeepPFCN consistently outperforms the other models on all the datasets, and each component contributed to the final result. DeepPFCN and MLFN shows the benefit of adding multi-scale architecture. MLFN and MLFN Fusion emphasize the importance of including the latent factor descriptor FS. MLFN-Fusion and ResNeXt display the priority of dynamic module selection.

\section{Conclusion}
\label{sec:conclusion}

We have proposed a simple but effective Deep Parallel Feature Consensus Net (DeepPFCN) approach for person re-identification. Experimental results with comparisons to other representative methods are provided, which indicate that the proposed approach outperforms other ensemble based person re-identification approaches, and achieves better than the state-of-the-art performance. Ablation study validates combination of modified MLFN with multi-scale consensus learning on improving performance.

\bibliographystyle{unsrt}  
\bibliography{references.bib}  



\end{document}